\documentclass{article}
\pdfoutput=1

\usepackage[margin=3cm]{geometry}
\usepackage[numbers]{natbib}

\usepackage{longtable}
\usepackage{booktabs}
\usepackage{siunitx}

\usepackage{graphicx}
\usepackage{algorithm2e}
\usepackage{subfigure}

\title{Robustness Evaluation of Regression Tasks with Skewed Domain Preferences}

\author{Nuno Costa\\ INESC TEC\\ Porto, Portugal \and Nuno Moniz\\Lucy Family Institute for Data \& Society\\ University of Notre Dame\\ Indiana, USA}

\date{}

\begin{document}

\maketitle


\begin{abstract}
In natural phenomena, data distributions often deviate from normality. One can think of cataclysms as a self-explanatory example: events that occur almost never, and at the same time are many standard deviations away from the common outcome. In many scientific contexts it is exactly these tail events that researchers are most interested in anticipating, so that adequate measures can be taken to prevent or attenuate a major impact on society. Despite such efforts, we have yet to provide definite answers to crucial issues in evaluating predictive solutions in domains such as weather, pollution, health. In this paper, we deal with two encapsulated problems simultaneously. First, assessing the performance of regression models when non-uniform preferences apply - not all values are equally relevant concerning the accuracy of their prediction, and there's a particular interest in the most extreme values. Second, assessing the robustness of models when dealing with uncertainty regarding the actual underlying distribution of values relevant for such problems. We show how different levels of relevance associated with target values may impact experimental conclusions, and demonstrate the practical utility of the proposed methods.
\end{abstract}

\section{INTRODUCTION}
Among the most common assumptions in statistical-based learning is the premise of normality. However, real-world domains commonly present departures (even if slightly) from normality, raising two critical caveats for machine learning. First, cases framed within the central tendency of the distributions are the most well-represented in the data. Thus, resulting models will tend to be better at recognising and anticipating such cases. Second, cases with extreme/rare values are often associated with high-impact events, e.g. fraud. Therefore, in certain situations, such cases should be prioritised and not under-valued w.r.t the ability to anticipate them.

Imbalanced learning is a sub-topic of machine learning that has gathered increasing attention for over 30 years. It mostly focuses on the ability to correctly model and predict rare and extreme events. Well-known as a very challenging machine learning task~\cite{Crone2005}, it usually faces issues hampering the learning process, including identifying small disjunct areas, lack of density and information in training data, overlapping classes, noisy data and data set shift (different distribution for training and test data)
~\cite{Lopez2013}. Since its inception, most of the research focus in this area concerns classification tasks, and work concerning regression is residual. In spite of this, the problem of imbalanced regression has witnessed an increasing level of interest in recent years. Such interest resulted in multiple contributions concerning the formalisation of the task in different data scenarios~\cite{Moniz2017a,oliveira2021biased} and multiple proposals for pre-processing strategies~\cite{torgo2015resampling,Moniz2018a}. However, the major linchpin for the challenge of imbalanced regression still lies on the issue of evaluation, and the need to adopt suitable metrics to correctly describe the performance of models in predicting extreme values. 

Work concerning the evaluation of imbalanced regression tasks is still deserving of much more attention. In respect to this, two major milestones include the work on utility-based metrics by Ribeiro~\cite{Ribeiro2011} -- based on adaptations of the popular F-Score~\cite{Rijsbergen1979} metric -- which were later reviewed by Branco~\cite{Branco2014} and Moniz~\cite{Moniz2017a}. This work has two main shortcomings. First, it is a threshold-based metric, in scenarios where a decision concerning which threshold to use is very sensitive and difficult to pinpoint. Second, the use of this threshold leads to the complete disregard of model performance on the majority of cases -- those deemed as non-extreme. More recently, Ribeiro and Moniz~\cite{Ribeiro2020a} proposed the Squared-Error Relevance Area (SERA) measure, which unlike previous work, disregards the use of any threshold while still focusing its evaluation of model performance on the more extreme values of the domain. Nonetheless, there is an issue that runs through both lines of previous work, which is that of robustness and the practical limitations of the foundations of imbalanced regression tasks. 

The most crucial concept in imbalanced regression tasks today concerns relevance functions~\cite{Torgo2007}. For now, it should suffice to clarify that it serves as a mapping between domain values, i.e. target values, and its importance (usually measured between 0 and 1) to the user in terms of obtaining an accurate prediction. Conceptually, this should be a user-defined process, at least partially -- it is unfeasible to associate a relevance judgement to all (infinite) values of a continuous domain. In such a scenario, users in possession of domain knowledge or experience would make judgements on the relevance of certain target values, and the remaining would be inferred, i.e. interpolated. However, in most situations, we lack such domain knowledge, and thus rely on automatic methods that are distribution-based. Essentially, those in the centre of the distribution would be associated to relevance judgements near 0, and as those values separate from such centre, their relevance should approximate (or assume the value of) 1. Naturally, this assumes that the available data describes a good sample of the underlying data domain, which is not always the case. Notwithstanding, even in the situation where some domain knowledge is available, small deviations from a ground truth might have a significant impact in model evaluation and selection, e.g. the lack of consensus in the quantification of biomarker concentration for characterisation of neuro-degenerative disease stages~\citep{chen2019changes}.

\subsection{Contributions}

In this paper, we present methods to ascertain the robustness of machine learning algorithms under domain preferences uncertainty. They are named convolution and elastic, and both consist in simulating a multitude of relevance scenarios through simulation, with the aim to understand how different algorithms perform under different relevance criteria. Each of these parts are associated to significant contributions, detailed as follows.

\begin{enumerate}
    \item \textbf{Metrics suitable for imbalanced numerical prediction.} Given that we assume there is greater relevance in identifying specific intervals of the target value, we cannot formalise prediction tasks as standard regression. Instead, we leverage recent work concerning imbalanced regression~\citep{Ribeiro2020a}, and the use of non-standard evaluation/optimisation metrics;
    \item \textbf{Robustness Evaluation.} Given the lack of consensus on what target values describe each regime of the underlying phenomena, we must evaluate the robustness of our models. We propose a robustness evaluation method that explores different ground-assumptions concerning the relation between target level regions and importance regimes.
\end{enumerate}

The remainder of the paper is organised as follows. Section~\ref{sec:background} describes relevant previous work and further contextualises the contributions of imbalanced regression in machine learning. The problem of imbalanced regression is detailed in Section~\ref{sec:imbreg}.
The robustness evaluation techniques are thoroughly explained in Section~\ref{sec:sweepstretch}.
Experimental results are thoroughly presented and discussed in Section~\ref{sec:expres}, followed by conclusions and future work in Section~\ref{sec:conclusions}.

\section{BACKGROUND}\label{sec:background}

The main assumption of standard metrics for evaluation of numerical prediction tasks is that each value in the domain has equal importance, i.e. uniform domain preferences. Several metrics have been proposed for learning settings where non-uniform preferences apply, i.e. a different level of relevance is associated to distinct ranges of the target variable. These range from asymmetrical approaches, common in the scope of finance, e.g. LIN-LIN~\citep{Christoffersen1996}, to adaptations of well-known metrics for classification tasks, such as the utility-based F-Score~\citep{Ribeiro2011,Branco2014,Moniz2017}. More recently, \citet{Ribeiro2020a} provide a thorough discussion on the limitations of existing metrics in the scope of learning with non-uniform preferences, and propose the Squared Error-Relevance Area (SERA) metric, which tackles several shortcomings in previous work -- we provide further details in Section~\ref{sec:imbreg}.

\subsection{The biomarker concentration problem - use case in detail}

\begin{figure}[!h]
    \centering
    \includegraphics[scale=0.5]{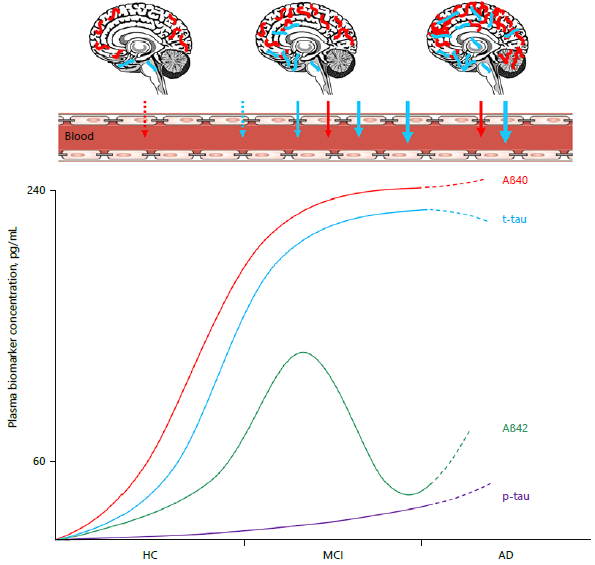}
    \caption{Peptide concentration across AD stages, from~\cite{chen2019changes}, concerning Healthy (HC), Mild cognitive impairment (MCI) and Alzheimer's (AD) stages.}
    \label{fig:peptide1}
\end{figure}

Medical researchers have been working on building concentration curves for different plasma biomarkers across the various stages of AD (Alzheimer's), and there are hints pointing to dynamic, nonlinear and nonparallel patterns over the course of the disease's development in an individual. Research has also shown there is a significant correlation between blood plasma and CSF biomarker concentration~\citep{cullen2021individualized}, a finding that reinforces the usefulness of blood-based biomarker tests. Notwithstanding, this can also be seen as an additional step of uncertainty in asserting the concentration values that require clinical intervention. In other words, concentration distributions for CSF and brain samples (where it's possible to find a ground truth) are still not well determined. Concentration distributions for blood plasma can only be inferred from the former - and it is virtually impossible to know the ground truth. In Figure~\ref{fig:peptide1}, it is possible to observe hypothetical concentration curves from 4 of the most popular AD biomarkers in plasma~\citep{chen2019changes}. 

This is a classic example where the underlying problem has a great scientific value and its solution would present a meaningful impact for neurodegenerative disease treatment. Unfortunately the scientific community is yet to unravel a precise regime classification through biomarker concentration levels.
It would, however, be possible to build a tentative relevance function using the information presented in Figure~\ref{fig:peptide1}.

\section{IMBALANCED REGRESSION}\label{sec:imbreg}

Standard regression tasks assume that a function $f()$ maps predictor variables to a continuous target variable. Such function is formalised as $Y = f(X_1, X_2, \cdots, X_p)$, where $Y$ is a numerical target variable, and features describing each case are denoted with $X_1, X_2, \cdots, X_p$. An approximation $h()$ (a model) to this function is obtained by using a data set with examples of the function mapping (known as a training set), i.e. $\mathcal{D}=\{\langle \mathbf{x}_i, y_i\rangle\}_{i=1}^n$. Here, our objective is to optimise models using a certain loss function, $L(\hat{y},y)$, such as the absolute or the squared error of estimations w.r.t. the true values. When all target values are considered equally important, i.e. uniform domain preferences, standard regression tasks assume that the utility $U$ of the estimations is a function of the estimation error $L(\hat{y},y)$ and that it is inversely proportional to the loss function, $U \propto L^{-1}$. However, this is not the case for learning tasks that are confounded by varying user preferences across range(s) of target values or in scenarios where estimation error may not hold as a useful metric across all possible target values. The aforementioned is the problem statement for imbalanced regression tasks.

\subsection{Relevance Functions}\label{subsubsec:relfunc}

\begin{figure}[ht!]
    \centering
    \includegraphics[scale=0.55]{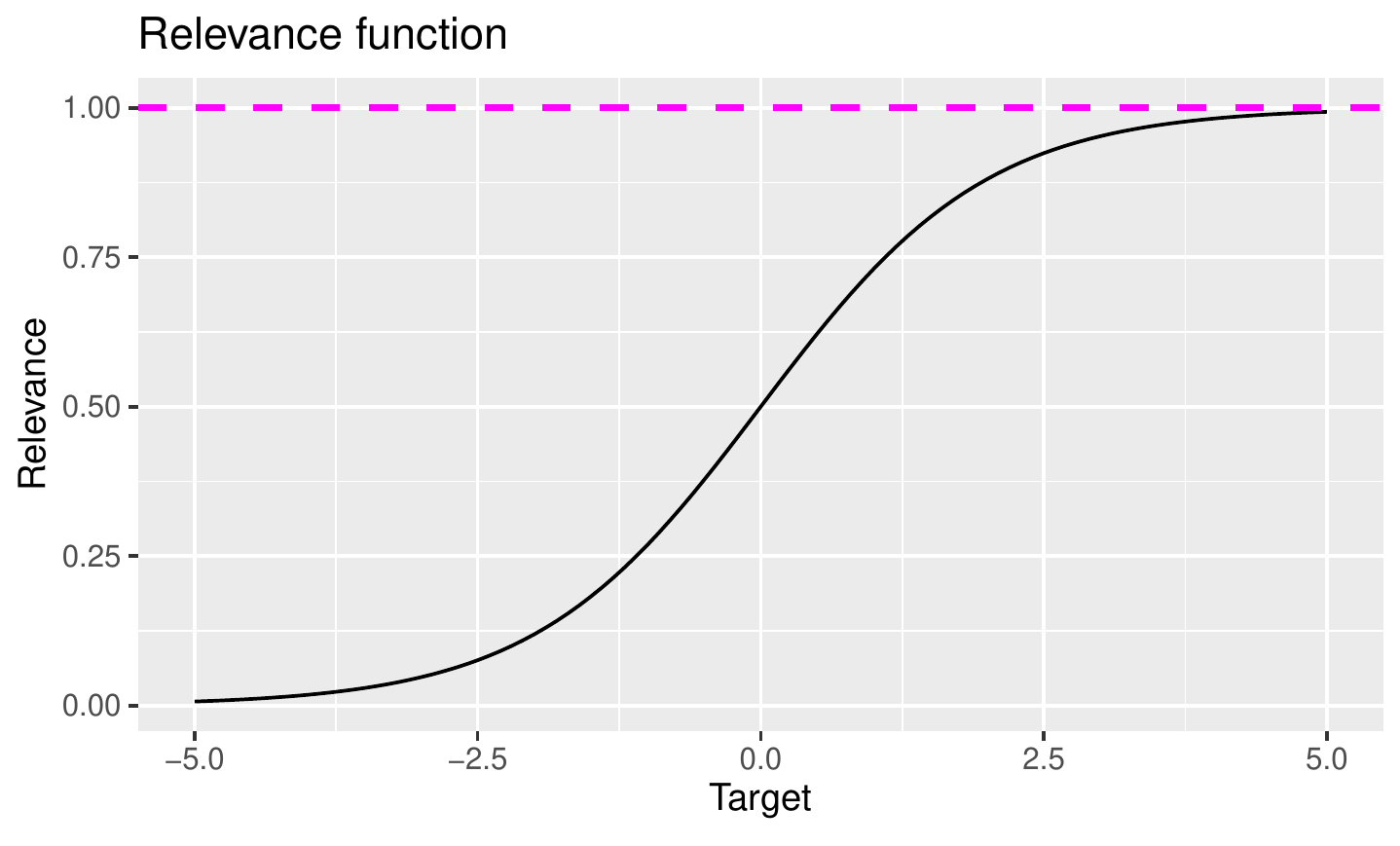}
    \caption{Example of target values' relevance for a specific domain preference. The dashed magenta line represents the case where no preferences are given, therefore all values are considered equally important.}
    \label{fig:imbalance}
\end{figure}

Imbalanced regression faces the challenge of providing a formal
approach capable of describing non-uniform preferences over continuous domains. Figure~\ref{fig:imbalance} illustrates the difference between the standard assumption of uniform vs.
non-uniform preferences. 

Ideally, the specification of domain preferences should be carried out by users or domain experts. Nonetheless, as previously discussed, this is often difficult (if not unfeasible). In such cases, we can rely of distribution-based statistics, as to automatically infer a suitable relevance function.

\paragraph{Definition} A relevance function $\phi(Y) : Y \to [0, 1]$ is a continuous function that encodes a user-defined domain preference Y by mapping it into a $[0, 1]$ scale of relevance, where 0 and 1 represent the minimum and maximum relevance, respectively~\citep{Ribeiro2020a}.

The relevance function construction starts with control points mapping to regions where relevance is known, ideally provided by domain experts. These are either specified by users or automatically derived from the data. When the latter is carried out, such process relies on the use of the adjusted box-plot method~\cite{Rousseeuw2011}, as suggested by~\citet{Ribeiro2020a}. From the control points, an interpolation method is used to approximate a continuous function ready to use in potentially infinite domains. For this occasion we will use Piecewise Cubic Hermite Interpolating Polynomials~\citep{dougherty1989nonnegativity}, also as suggested by~\citet{Ribeiro2011}.

\subsection{Evaluation}\label{subsubsec:eval}

Concerning the evaluation of solutions in imbalanced regression tasks, we propose the use of a recent evaluation metric, SERA (Squared Error-Relevance Area), proposed by~\citet{Ribeiro2020a}. Such a metric is able to enhance the quantitative focus in ranges other than the most common values, leveraging the previously presented concept of relevance functions.

\paragraph{Definition} For a given data set D and a relevance function $\phi$, consider the subset $D^t$ where the relevance of the target value is above or equal to a cutoff $t$. In such case, it's possible to obtain an estimate of the Squared Error-Relevance of a model as: 

\begin{equation}
   SER_t =  \sum_{i\in D^t }( \widehat{y}_i - y_i)^2
\end{equation}

where $\widehat{y}_i$ and $y_i$ are the estimated and true values for data point $i$, respectively. For this estimate, only the subset of estimations for which the relevance of the true target value is above a known threshold t, are considered.
SERA is defined as:

\begin{equation}
   SERA =  \intop\nolimits_{0}^{1} SER_tdt= \intop\nolimits_{0}^{1} \sum_{i\in D^t }( \widehat{y}_i - y_i)^2dt
\end{equation}

In general terms, the smaller the area under the curve (SERA), the better is the model's performance. In a situation where no domain preferences are defined, i.e. $\phi(Y) = 1$ , SERA converges to the sum of squared errors. For a more in-depth analysis and presentation of this recent proposal, we recommend the reading of~\cite{Ribeiro2020a}.

\section{EVALUATION OF ROBUSTNESS UNDER UNCERTAINTY}\label{sec:sweepstretch}

The formal definition for imbalanced regression tasks provided by~\citet{Ribeiro2020a} provides the basis for tackling the first issue in this project: how to evaluate and compare solutions for which the relevance of obtaining correct estimations in different target values is different? Indeed, having a more appropriate metric for the evaluation of regression models on extreme target values, we are left with the second challenge to ponder over: in situations
of uncertainty concerning the exact value range of predictive focus, what is the best model to ensure a robust solution? 

In this section, we propose two methods that can be applied in tandem with a SERA-based evaluation of predictive solutions. Such methods are based on the simulation of neighbouring configurations of the relevance function based on the initial hint. The overall intuition is that by evaluating models in a considerable amount of near configurations (w.r.t the original ``hint'') for SERA, it is possible to understand the behaviour of different algorithms/strategies under uncertainty. It is also possible to understand what is the most robust method under such events. To derive functions in the vicinity of the relevance function constructed from the original data, we propose two methods to simulate two degrees of freedom: \textit{i)} x-axis displacement and \textit{ii)} slope incline. For both methods, the variation range can be established according to the problem needs.

\begin{itemize}
    \item \textbf{Convolution.} First (Algorithm~\ref{alg:conv}), we can think of the situation in which the hint regarding the shape is correct, but inaccurate concerning which are the most relevant values. By shifting the relevance function across a certain range -- much like a convolution -- it is possible to collect the performance of models for an additional number of similarly derived relevance functions;
    \item \textbf{Elastic.} Second (Algorithm~\ref{alg:elastic}), we consider a situation where the relevance function derived from the original data is correct about which values are the most relevant, but not correct about the shape. It consists in fixating the point where the relevance is larger than 0, while varying the first point where the relevance is equal to 1.
\end{itemize}

For both methods we start by determining the range of the target variable $Y \in \mathcal{Y}$ and obtaining a relevance function $\phi(Y)$. The function should be a mapping of the domain $\mathcal{Y}$ into a $[0,1]$ scale of relevance and it should be a smooth function. Next, we define the peak relevance $\pi_{max}$ for $\phi(Y)$, as the minimum value for which $\phi(Y)=1$, i.e. the first maxima encountered. Similarly, we define the base relevance, $\pi_{min}$ as the largest target value for which $\phi(Y)=0$, i.e. the point immediately before the slope of $\phi(Y)$. Note that, within the scope of our work, we are dealing with one-tail right distribution of values. In case of a distribution skewed left, the peak and base relevance values are exchanged.

\begin{algorithm}[!h]
\SetAlgoLined
\KwResult{Performance for Convolution method}
 $Y =$ training set target variable\; 
 $\delta =$ search step size\; 
 $\phi()= $initial relevance function\;
 $\pi_{max}= $initial peak relevance\;
 $\pi_{min}= $initial base relevance\;
 $\textbf{n}= \frac{range(Y)}{\delta}$\;
 \For{i in \textbf{n}}{
  update relevance function $\phi()$ s.t. $\pi_{max}=\pi_{max}+\delta$ and $\pi_{min}=\pi_{min}+\delta$\;
  evaluate the model using SERA with $\phi()$ as the criterion\;
 }
 \caption{Convolution simulation for relevance function}\label{alg:conv}
\end{algorithm}

\begin{algorithm}[!h]
\SetAlgoLined
\KwResult{Performance for Elastic method}
 $Y =$ training set target variable\;
 $\delta =$ search step size\; 
 $\phi() = $initial relevance function\;
 $\pi_{max} = $initial peak relevance\;
 $\pi_{min} = $initial base relevance\;
 M = $\tilde{Y}$, median $Y$\;
 $\textbf{n} = \frac{range(M,M+\sigma(Y))}{\delta}$\;
 \For{i in \textbf{n}}{
  update relevance function $\phi()$ s.t. $\pi_{max}=\pi_{max}+\delta$ and $\pi_{min}=\pi_{min}$\;
  evaluate the model using SERA with $\phi()$ as the criterion\;
}
\caption{Elastic simulation for relevance functions.}\label{alg:elastic}
\end{algorithm}

\begin{figure}[ht!]
    \centering
    \includegraphics[width=0.5\textwidth]{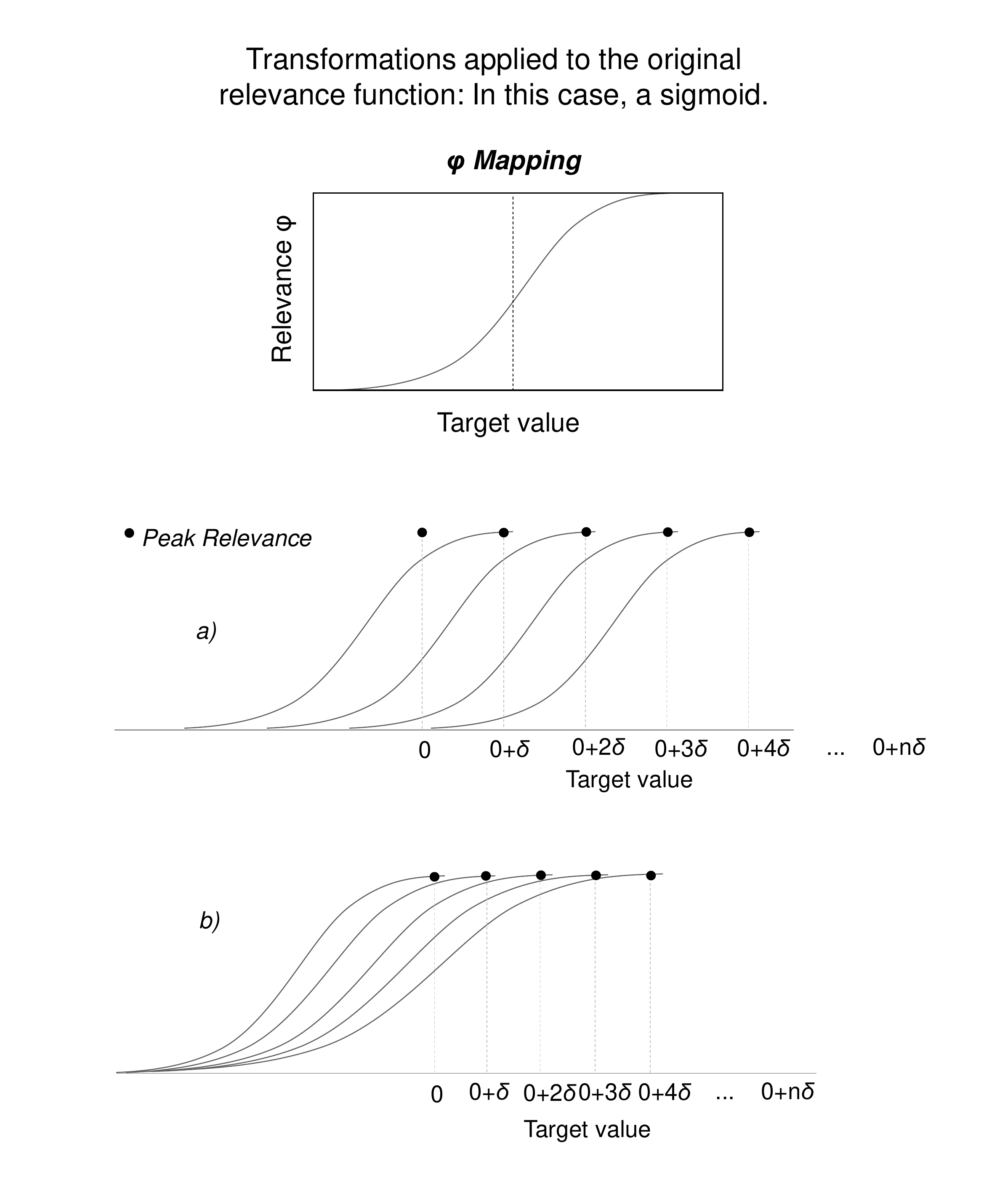}
    \caption{Left - Original relevance function. Dashed line is the inflection point. Right - Relevance methods: a) Convolution b) Elastic}
    \label{fig:relevancescheme} 
\end{figure}

Figure~\ref{fig:relevancescheme} illustrates the transformations performed on a sigmoid relevance function: a) refers to Convolution, while b) refers to Elastic.

In closing, we should note the following. For our experimental study, we use a Convolution range from $\pi_{max}$-$\sigma_{target}$ up to $\pi_{max}$+$\sigma_{target}$ of the target value, while the Elastic
range we define as $\pi_{1}$-$\sigma_{target}$ up to $\pi_{1}$+$\sigma_{target}$ for the first extreme $\pi,$ and $\pi_{2}$+$\sigma_{target}$ up to $\pi_{1}$-$\sigma_{target}$ for the second extreme $\pi$. Extremes can be $\phi=1$ or $\phi=0$

\section{EXPERIMENTAL STUDY}\label{sec:expres}

In this section, we present an experimental study concerning the usefulness of our robustness assessment approaches using multiple data sets. We start by presenting the main research questions that drive our study: how robust are the conclusions concerning the performance of predictive models in the context of relevance uncertainty?

\subsection{Methods}

In this subsection, we provide experimental details concerning data, including adopted methodology, custom feature selection methods, learning algorithms and estimation procedures.

\subsubsection{Learning Algorithms}

For this study, we consider the following algorithms: Multivariate Adaptive Regression Splines (\textit{mars}), Extreme Gradient Boosting (\textit{xgboost}) and Random Forest. These are listed in Table~\ref{tab:hyperparams}, with the respective hyperparameter grid and implementation details. The hyperparameters pertaining to \textit{xgboost} and Random Forest parameters are chosen using a random search. From the high-dimensional grid space, 15 random combinations are considered. On the other hand, MARS evaluates 24 fixed combinations.

\begin{table}[!h]
    \centering
    \scriptsize
    \begin{tabular}{llll}  
    \toprule
    Algorithm&Grid size&Parameters&R Package\\
    \midrule
    \hline
    
    mars & 24 & degree $\in \{1,2,3\}$ & earth\\ 
    & & nprune $\in \{1,2,3,4,6,7,8\}$ &\\
    \hline
    xgboost & 50 & nrounds $\in \{1:1000\}$ & xgboost\\
    & & max depth $\in \{1:10\}$ & \\
    & & eta $\in \{0,001:0,6\}$ & \\
    & & gamma $\in \{0:10\}$ & \\
    \hline
    random forest & 50 & mtry $\in \{0:92\}$ & ranger\\
    & & min node size $\in \{1,2,3,4,5\}$ & \\
    & & n.trees = 500 & \\
    \hline
  \bottomrule
\end{tabular}
    \caption{Parameters considered for grid search during model training}\label{tab:hyperparams}
\end{table}

\subsubsection{Estimation}\label{subsubsec:estimation}

\paragraph{Validation.} Concerning estimation methodologies for the purposes of validation, we adopt a 10-fold cross validation method~\citep{Kohavi1995} on the train set for hyperparameter optimisation. 

\paragraph{Relevance function.} The data sets used in this study are of unknown nature, apart from features' names. We are not capable of introducing any domain expertise to build any kind of guess to what the relevance profile should look like. Therefore, we assume the relevance function automatically generated according to \citet{Ribeiro2020a} as our reference.

\begin{figure*}[!ht]
    \centering
    \subfigure[Germany data set]{\includegraphics[width=0.32\textwidth]{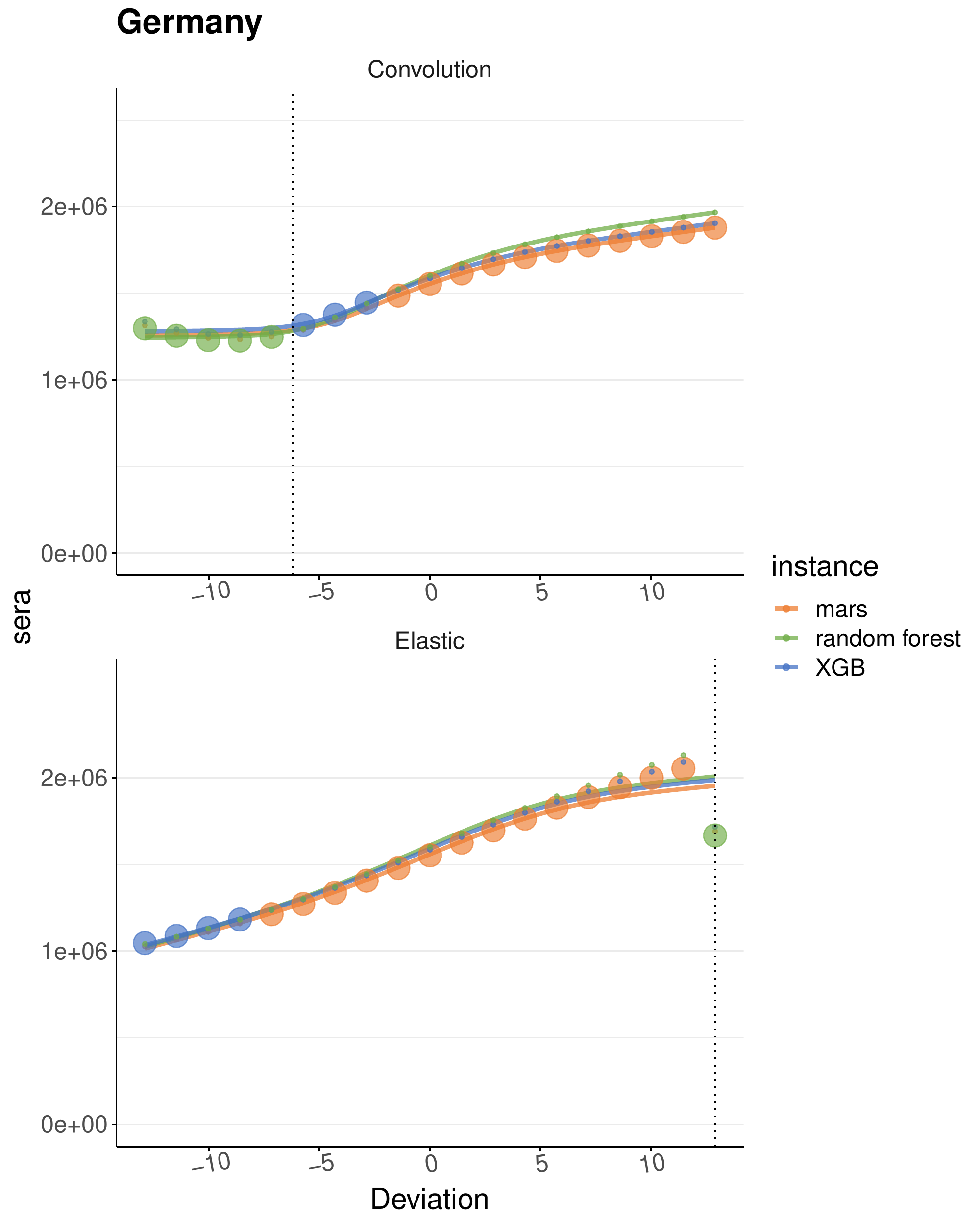}} 
    \subfigure[Beijing data set]{\includegraphics[width=0.32\textwidth]{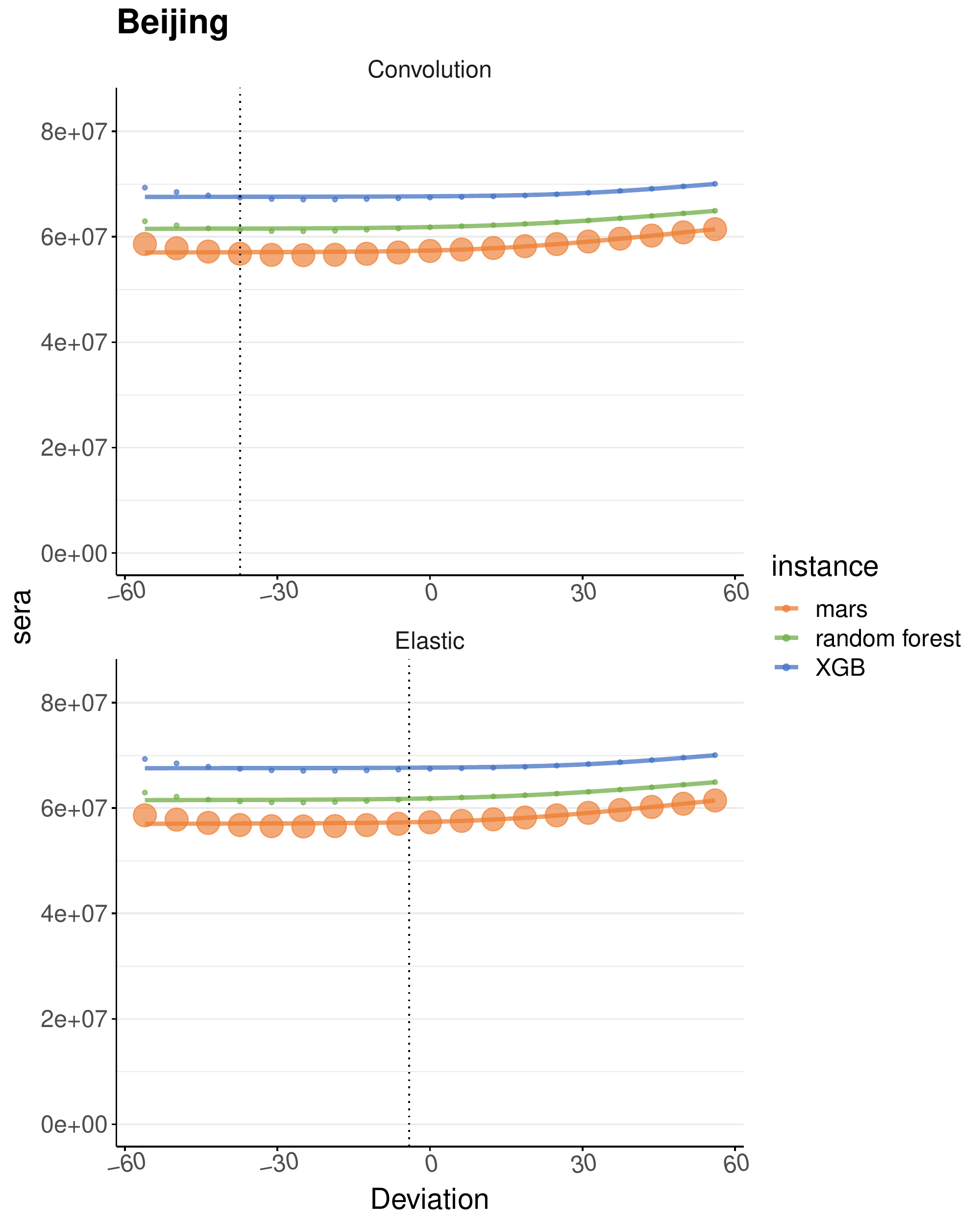}} 
    \subfigure[Norway data set]{\includegraphics[width=0.32\textwidth]{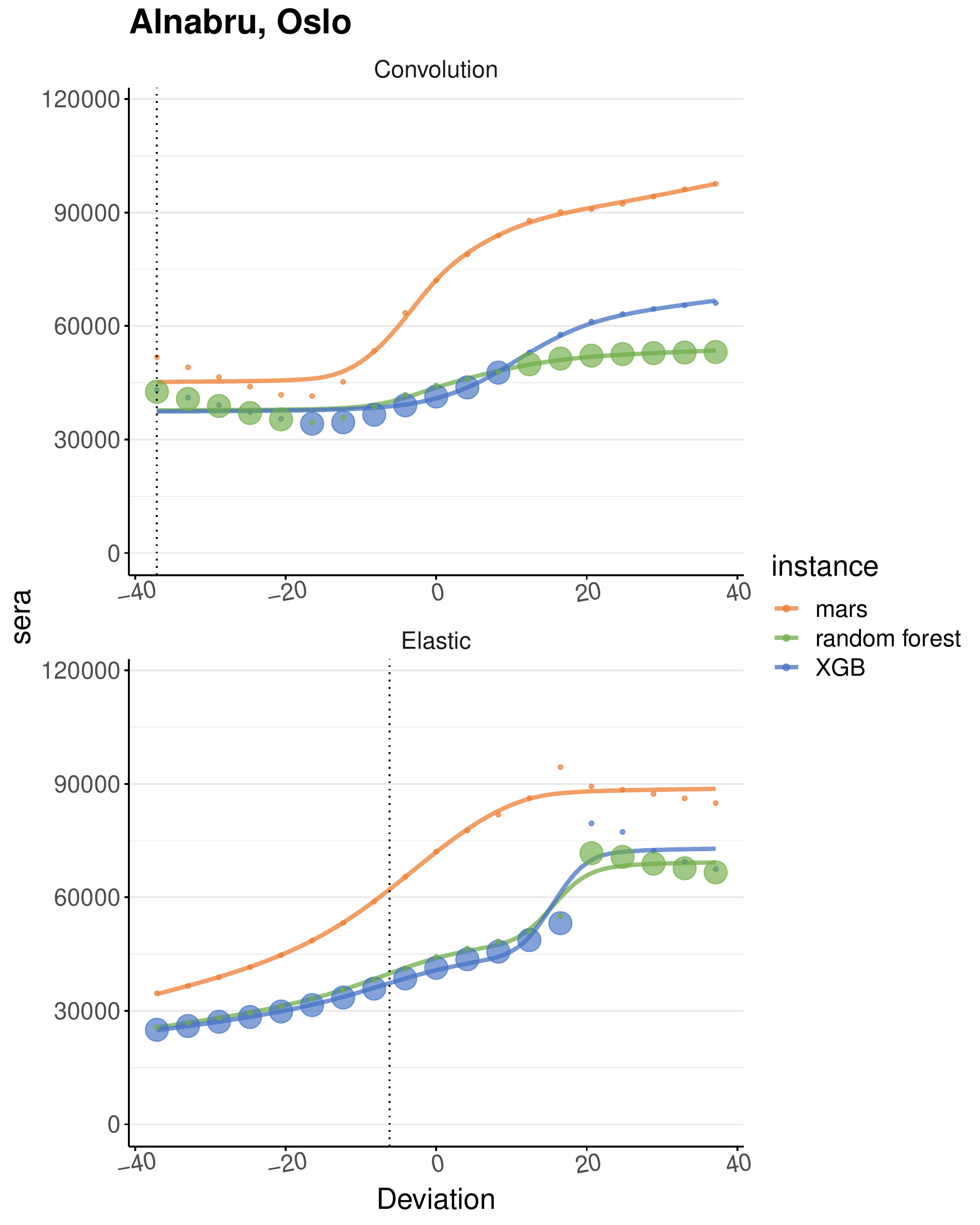}}
    \caption{Pollution data sets. In each chart, the vertical dashed line corresponds to the nearest approximation of the domain knowledge relevance function.}\label{fig:countryfigs}
\end{figure*}

For the Convolution technique, we propose to simulate neighbouring configurations of relevance functions within the range of one standard deviation to the left and right of the automatically generated points. For example, for a function with peak relevance at 100, i.e. $\phi(100)=1$, we will simulate scenarios where values between $100-\sigma$ and $100+\sigma$ will be reference points for peak relevance. In short, the relevance function is incrementally shifted to both sides in its original form. For the Elastic technique, we apply it in the same range, but this time moving the left extremity further to the left while the right extremity moves to the right, and vice-versa. A good analogy is as if one decides to stretch and shrink the original relevance function by pulling/pushing its peaks (for 1 and 0)
Of course, this is an educated guess of different scenario assumptions, enabling to explore the data. We again emphasise that our final goal is to find a method that can better withstand such uncertainties regarding relevance bands.

\paragraph{Optimisation criteria.} As previously discussed, a particularity of our learning setting is that, differently from the standard setting of regression tasks, our work is set in a context where one assumes that different values have different relevance to users. Previous work has extensively demonstrated how the use of traditional optimisation criteria within this scope may be misleading~\citep{Ribeiro2020a}. Accordingly, our analysis concerning which are the best hyper-parameter combinations within the scope of this experimental study is based on the use of the SERA metric (Section~\ref{subsubsec:eval}).

\subsection{Results}

In this section, our goal is to demonstrate that the techniques proposed in this paper are capable of properly assessing the robustness of predictive models' performance in the context of uncertainty in imbalanced regression tasks. Accordingly, we divide this section in two parts:

\begin{enumerate}
    \item Case-study analysis of the Convolution and Elastic techniques applied to three air pollution data sets and the indicator $PM10$, to grasp the impact of using such techniques with policy-informed control points;
    \item Evaluation of prediction models in all available data sets concerning the robustness of their performance when applying our our proposed techniques;
\end{enumerate}

\begin{figure*}[!ht]
    \centering
    \includegraphics[width=\textwidth]{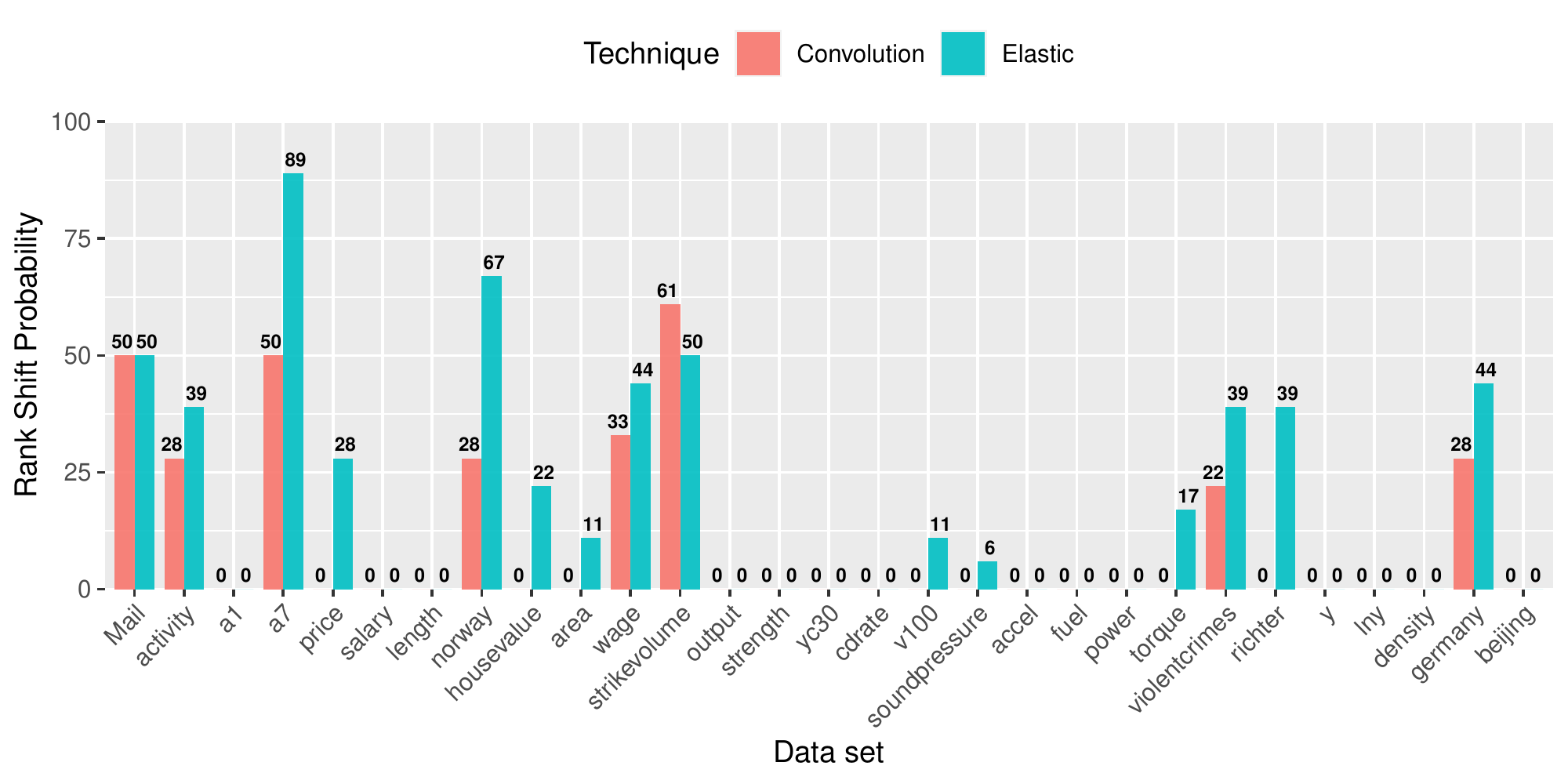}
    \caption{Rank shift probability for all data sets according to the application of the Convolution and Elastic techniques.}
    \label{fig:rankshiftprob} 
\end{figure*}

Concerning the first part, a note on the data sets used. The first data set concerns hourly averages of concentration levels in Beijing (China) from August 2012 to March 2013~\cite{zheng2013u}; the second, daily averages for rural background stations in Germany from 1998 to 2009~\cite{pebesma2012spacetime}; and, finally, hourly values for a station in Alnabru, Oslo (Norway), between October 2001 and August 2003~\cite{aldrin2006improved}. The benefit of using these data sets is that we can base the set of control points used on official recommendations per the World Health Organisation~\cite{press2014air} for denoting 24-hour averages as normal or dangerous: $\phi(50\mu$g$/m)=0$ and $\phi(150\mu$g$/m)=1$, respectively. Figure~\ref{fig:countryfigs} illustrates the outcomes for the best models of each learning algorithm used, estimated according to the methodology described in Section~\ref{subsubsec:estimation}. The scenario corresponding to the x-axis value of $0$ concerns the relevance function using policy-informed control points.

A careful analysis of the results provides interesting insights concerning possible scenarios. For example, concerning the first data set (left), we observe that, when applying both techniques, there is a considerable performance overlap despite different underlying relevance functions. In the second data set (centre), we clearly notice a consistent distinction between each of these models throughout the entire set of neighbouring relevance functions. As for the third (right), we gain insights into how conclusions concerning a certain relevance function may be brittle. In this case, we find that small shifts in the reference points for relevance judgements may have great impact in model evaluation and comparison. It also illustrates how different shapes of the underlying relevance function may have distinct impacts -- by comparing the impact of the Convolution and the Elastic techniques.

\begin{figure*}[!ht]
    \centering
    \subfigure[Data set: Output]{\includegraphics[width=0.32\textwidth]{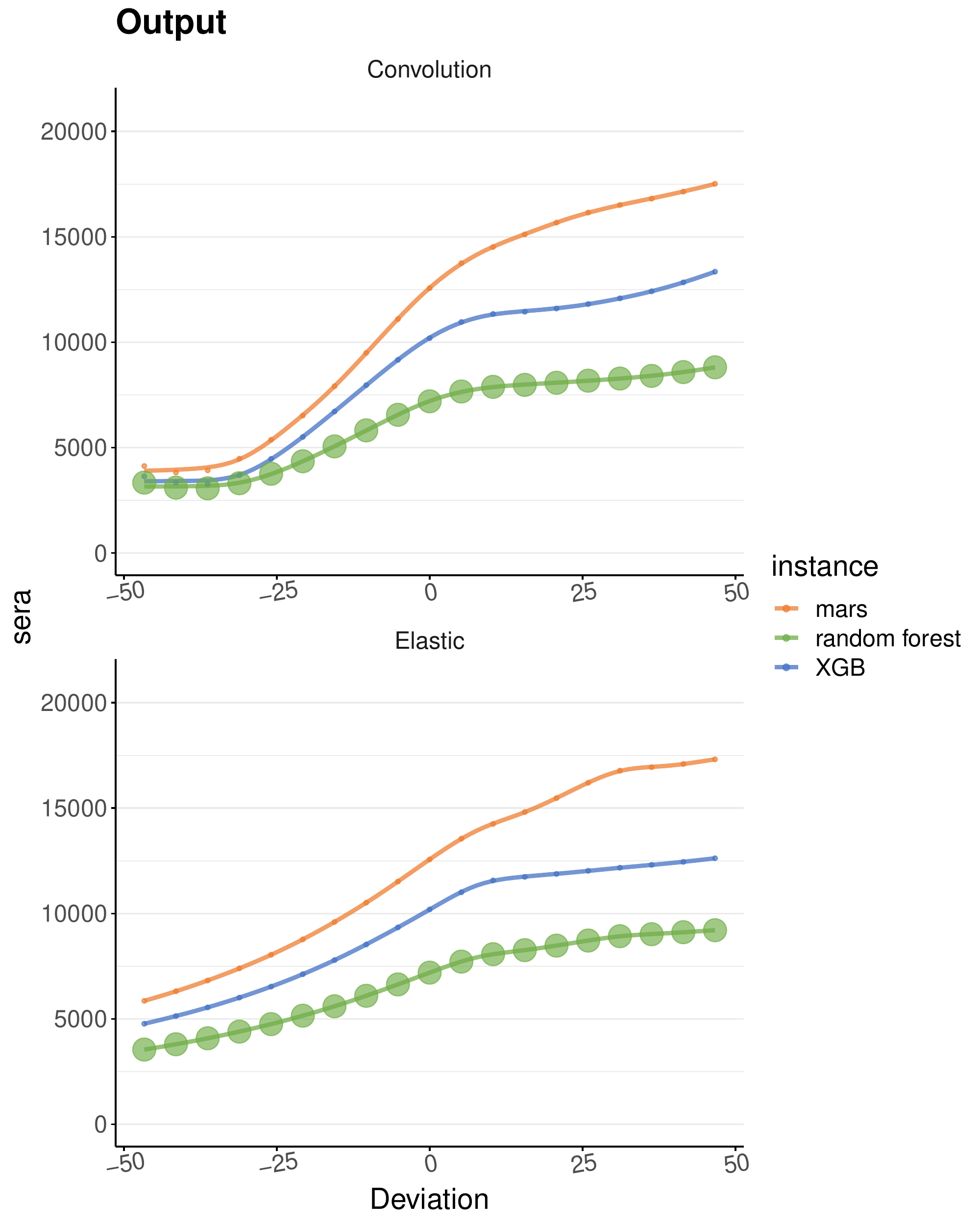}}
    \subfigure[Data set: Richter]{\includegraphics[width=0.32\textwidth]{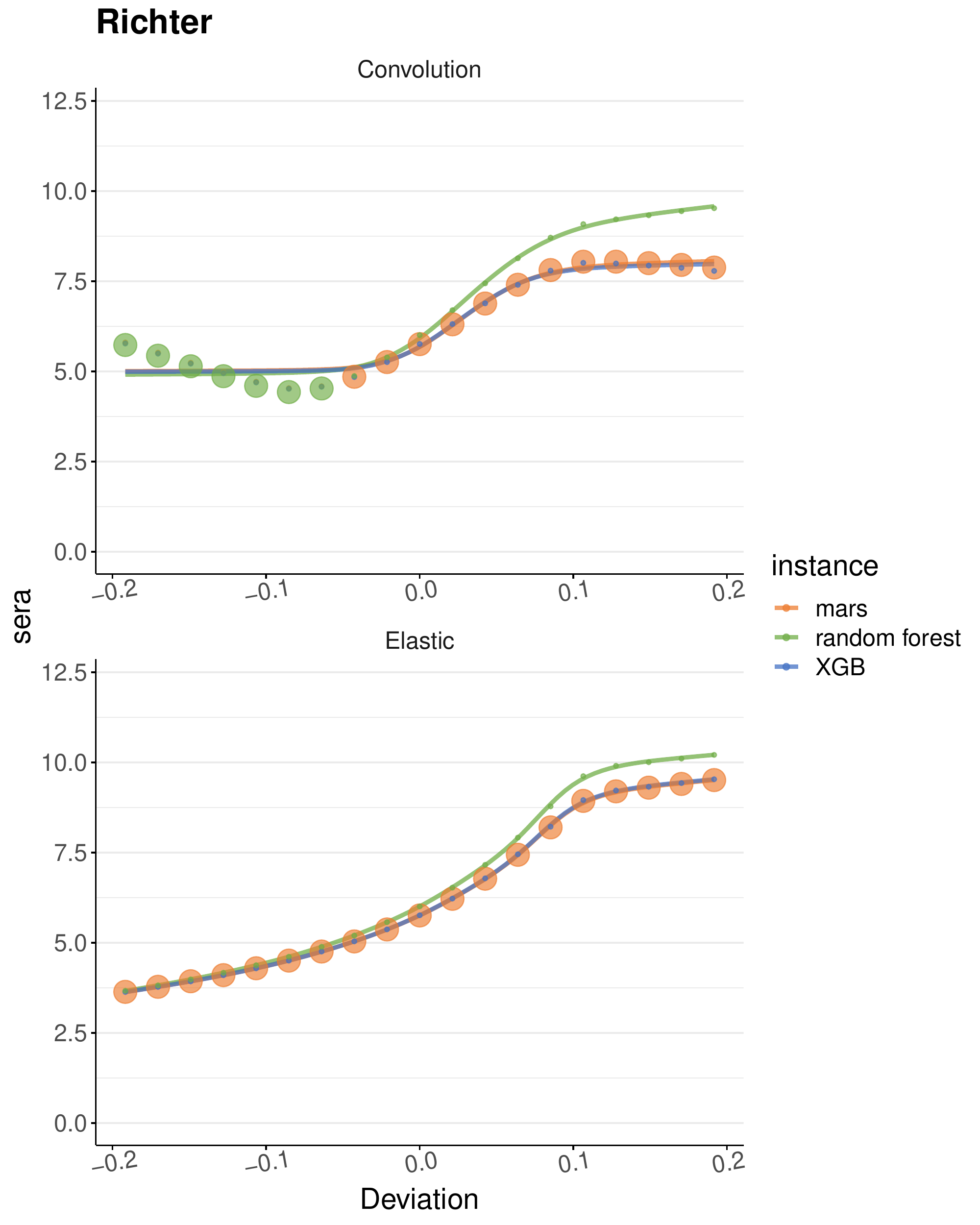}} 
    \subfigure[Data set: a7]{\includegraphics[width=0.32\textwidth]{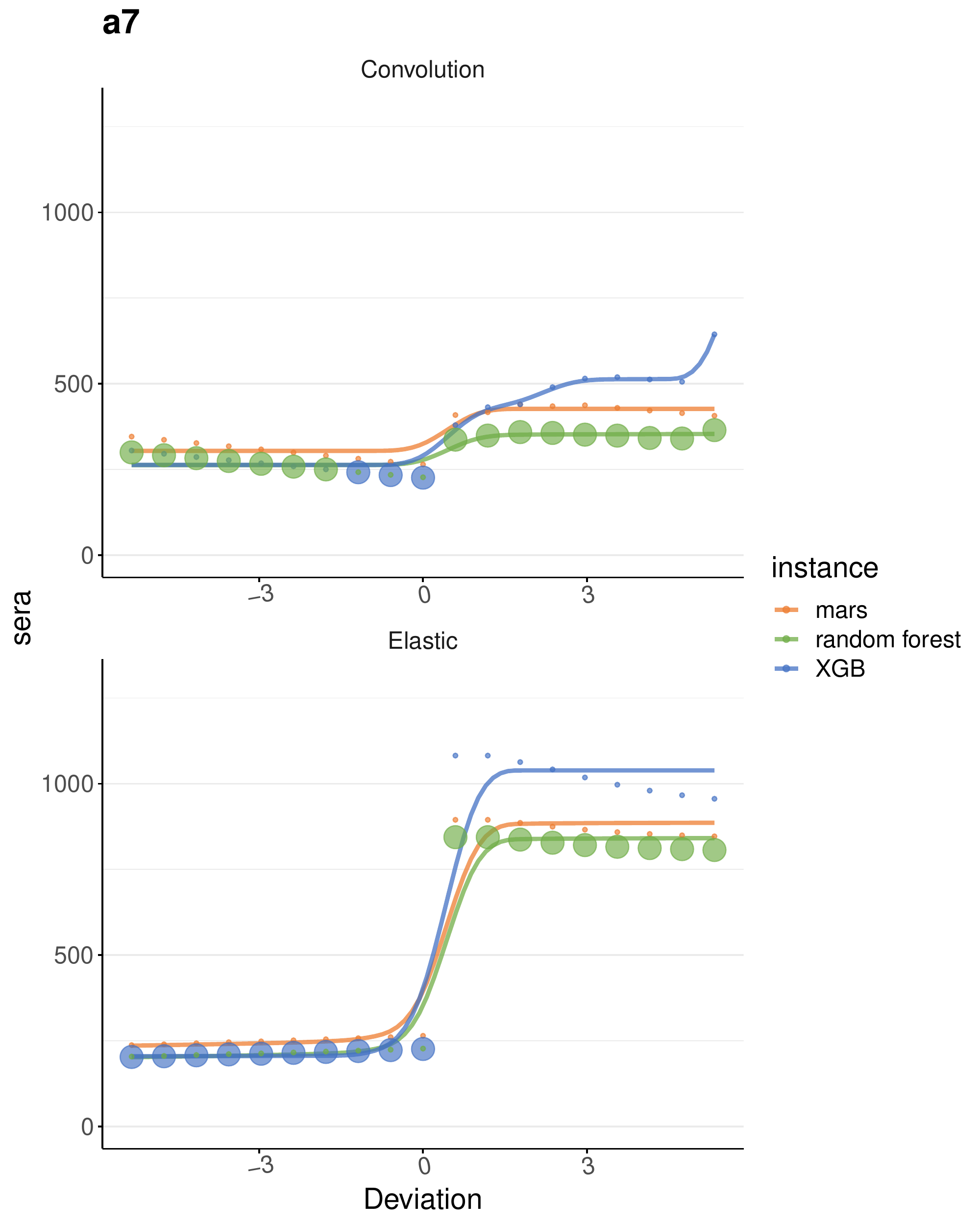}} 
    \caption{Performance of models in experimental data sets under the application of the Convolution and Elastic techniques}\label{fig:usecases}
\end{figure*}

Concerning the second part, we test the Convolution and Elastic techniques in the 29 data sets, having as reference the relevance function defined by the automatic method proposed in~\citet{Ribeiro2020a}. The experimental resolution for neighbouring configurations when applying our proposed techniques is 19 steps for Convolution and 19 steps for the Elastic technique. Results are presented in two distinct manners. First, illustrating an overview of the experimental results, Figure~\ref{fig:rankshiftprob} describes the rank shift probability for the prediction models estimated as the best (best rank) among all candidates. The rank shift probability translates as the percentage of neighbouring scenarios where the best model according to the automatic relevance function did not remain as the best. In other words, a robust scenario would describe a near 0\% of rank shift probability, meaning that the best model according to initial performance rank remained as such in all neighbouring scenarios tested, and vice-versa. Second, by analysing concrete cases that demonstrate the usefulness of using the proposed techniques, based on interesting scenarios from the first set of results -- see Figure~\ref{fig:usecases}.

As an overall assessment of the results -- see Figure~\ref{fig:rankshiftprob} -- we observe the same type of scenarios as in the first part of our experimental results. In almost half of the data sets used (14), neither of the proposed techniques demonstrated performance rank shifts for the best model w.r.t the reference relevance function. Around a quarter of the data sets demonstrated varying levels of rank shift probability when applying both the Convolution and Elastic techniques. We should highlight cases at both extremes, such as with the data sets \textit{a7} and \textit{violentcrimes}. The former presents probabilities of 50\% and 89\% for Convolution and Elastic techniques, respectively; and, for the latter, 17\% and 22\%. Finally, we observe 6 cases where only the application of the Elastic technique rendered a positive probability. No cases with such behaviour were observed when applying the Convolution technique. In this aspect, we should mention the case of the data set \textit{Richter}, with 39\% of rank shift probability for the Elastic technique and 0\% for the Convolution technique.

To better grasp the usefulness of the proposed techniques, we propose to look closer at the results (Figure~\ref{fig:usecases}) from three data sets and their illustrative scenarios: the \textit{Output}, \textit{Richter} and \textit{a7} data sets. In the \textit{Output} data set, we find a scenario of robustness in terms of performance, where simulations based on both techniques conclude that the best initial performer (in this case a Random Forest model) stays as such in all neighbouring configurations. As for the \textit{Richter} data set, we observe a good example where, if the relevance function is convoluted to the right, algorithms present a similar performance, but when the relevance function is convoluted to the left, the best model (from the \textit{mars} algorithm) is no longer the best option for the particular imbalanced regression task. Finally, regarding the \textit{a7} data set we observe a considerable discrepancy between results for both the Convolution and Elastic techniques when contracting/expanding the form of the relevance function or when sliding it right or left. This impact raises concrete issues on the particular sensitivity of some domains to small changes in assertions concerning what is most important to be accurate in terms of predictive performance.

Given the experimental results obtained concerning the proposed methods, we highlight the following conclusions.

\begin{itemize}
    \item \textbf{Regime shift.} Using both techniques, we observe that some data sets are sensible to shifts in the relevance criteria for the target value, leading to the best predictive model losing its initial advantage. 
    \item \textbf{Function form.} We also note that the duality between the approach in the Convolution and Elastic techniques is useful: in several cases, the Elastic technique demonstrates differences in model dominance, while the Convolution technique does not;
    \item \textbf{Variety of scenarios.} We consider a valuable contribution the diversity that both the Elastic and Convolution techniques introduce in trying to understand algorithm dominance under different relevance scenarios.
\end{itemize}

\section{CONCLUSIONS AND FUTURE WORK}\label{sec:conclusions}

In this paper, we propose two simulation-based techniques that allow key insights into the robustness of prediction models in imbalanced regression tasks. We note that this is a particularly common scenario where domain knowledge regarding which values are more important is not available. The two proposed methods, Convolution and Elastic, extend the more traditional performance evaluation process with the use of multiple relevance functions aiming at the simulation of different scenarios of target value's relevance regions. Altogether, these methods have allowed us to better evaluate each model's performance on the relevant target value band, and by extending the analysis to multiple relevance configurations, we are able to identify the most consistent and robust model, i.e. the ones that will consistently perform better than others regardless of which target value range is deemed the most relevant.

Based on the experimental analysis of these methods provided in this paper, we demonstrate how they can be useful for scientists working with imbalanced regression problems where domain knowledge is unavailable or even yet to answer details about which values are more important to predict -- a known scenario is key challenges such as early detection of Alzheimer's disease.

For the sake of reproducible science, all data sets and code necessary to reproduce the results presented in this paper are available in (removed for anonymous submission). 


\subsubsection*{References}
\bibliography{bibdb}

\begin{thebibliography}{}

\bibitem[Aldrin, 2006]{aldrin2006improved}
Aldrin, M. (2006).
\newblock Improved predictions penalizing both slope and curvature in additive
  models.
\newblock {\em Computational statistics \& data analysis}, 50(2):267--284.

\bibitem[Branco, 2014]{Branco2014}
Branco, P. (2014).
\newblock {\em Re-sampling Approaches for Regression Tasks under Imbalanced
  Domains}.
\newblock PhD thesis, Universidade do Porto.

\bibitem[Chen et~al., 2019]{chen2019changes}
Chen, T.-B., Lai, Y.-H., Ke, T.-L., Chen, J.-P., Lee, Y.-J., Lin, S.-Y., Lin,
  P.-C., Wang, P.-N., and Cheng, I.~H. (2019).
\newblock Changes in plasma amyloid and tau in a longitudinal study of normal
  aging, mild cognitive impairment, and alzheimer’s disease.
\newblock {\em Dementia and geriatric cognitive disorders}, 48(3-4):180--195.

\bibitem[Christoffersen and Diebold, 1996]{Christoffersen1996}
Christoffersen, P.~F. and Diebold, F.~X. (1996).
\newblock Further results on forecasting and model selection under asymmetric
  loss.
\newblock {\em Journal of Applied Econometrics}, 11(5):561--571.

\bibitem[Crone et~al., 2005]{Crone2005}
Crone, S.~F., Lessmann, S., and Stahlbock, R. (2005).
\newblock Utility based data mining for time series analysis: Cost-sensitive
  learning for neural network predictors.
\newblock In {\em Proceedings of the 1st International Workshop on
  Utility-based Data Mining}, UBDM '05, pages 59--68, New York, NY, USA. ACM.

\bibitem[Cullen et~al., 2021]{cullen2021individualized}
Cullen, N.~C., Leuzy, A., Palmqvist, S., Janelidze, S., Stomrud, E., Pesini,
  P., Sarasa, L., Allu{\'e}, J.~A., Proctor, N.~K., Zetterberg, H., et~al.
  (2021).
\newblock Individualized prognosis of cognitive decline and dementia in mild
  cognitive impairment based on plasma biomarker combinations.
\newblock {\em Nature Aging}, 1(1):114--123.

\bibitem[Dougherty et~al., 1989]{dougherty1989nonnegativity}
Dougherty, R.~L., Edelman, A.~S., and Hyman, J.~M. (1989).
\newblock Nonnegativity-, monotonicity-, or convexity-preserving cubic and
  quintic hermite interpolation.
\newblock {\em Mathematics of Computation}, 52(186):471--494.

\bibitem[Kohavi, 1995]{Kohavi1995}
Kohavi, R. (1995).
\newblock A study of cross-validation and bootstrap for accuracy estimation and
  model selection.
\newblock In {\em Proceedings of the 14th International Joint Conference on
  Artificial Intelligence - Volume 2}, IJCAI’95, page 1137–1143, San
  Francisco, CA, USA. Morgan Kaufmann Publishers Inc.

\bibitem[L{\'o}pez et~al., 2013]{Lopez2013}
L{\'o}pez, V., Fern{\'a}ndez, A., Garc{\'\i}a, S., Palade, V., and Herrera, F.
  (2013).
\newblock An insight into classification with imbalanced data: Empirical
  results and current trends on using data intrinsic characteristics.
\newblock {\em Information Sciences}, 250:113--141.

\bibitem[Moniz et~al., 2017a]{Moniz2017a}
Moniz, N., Branco, P., and Torgo, L. (2017a).
\newblock Evaluation of ensemble methods in imbalanced regression tasks.
\newblock In Torgo, L., Krawczyk, B., Branco, P., and Moniz, N., editors, {\em
  Proceedings of the First International Workshop on Learning with Imbalanced
  Domains: Theory and Applications}, volume~74 of {\em Proceedings of Machine
  Learning Research}, pages 129--140, ECML-PKDD, Skopje, Macedonia. PMLR.

\bibitem[{Moniz} et~al., 2018]{Moniz2018a}
{Moniz}, N., {Ribeiro}, R.~P., {Cerqueira}, V., and {Chawla}, N. (2018).
\newblock Smoteboost for regression: Improving the prediction of extreme
  values.
\newblock In {\em 2018 IEEE 5th Int. Conf. on Data Science and Advanced
  Analytics (DSAA)}, pages 150--159.

\bibitem[Moniz et~al., 2017b]{Moniz2017}
Moniz, N., Torgo, L., Eirinaki, M., and Branco, P. (2017b).
\newblock A framework for recommendation of highly popular news lacking social
  feedback.
\newblock {\em New Generation Computing}, 35(4):417--450.

\bibitem[Oliveira et~al., 2021]{oliveira2021biased}
Oliveira, M., Moniz, N., Torgo, L., and Santos~Costa, V. (2021).
\newblock Biased resampling strategies for imbalanced spatio-temporal
  forecasting.
\newblock {\em International Journal of Data Science and Analytics},
  12(3):205--228.

\bibitem[Pebesma, 2012]{pebesma2012spacetime}
Pebesma, E. (2012).
\newblock spacetime: Spatio-temporal data in r.
\newblock {\em Journal of statistical software}, 51:1--30.

\bibitem[Press, 2014]{press2014air}
Press, W. (2014).
\newblock Who air quality guidelines for particulate matter, ozone, nitrogen
  dioxide and sulfur dioxide.
\newblock {\em World Health Organization. http://whqlibdoc. who.
  int/hq/2006/WHO\_SDE\_ PHE\_OEH\_06. 02\_eng. pdf. Accessed}, 25.

\bibitem[Ribeiro, 2011]{Ribeiro2011}
Ribeiro, R. (2011).
\newblock {\em Utility-based Regression}.
\newblock PhD thesis, Dep. Computer Science, Faculty of Sciences - University
  of Porto.

\bibitem[Ribeiro and Moniz, 2020]{Ribeiro2020a}
Ribeiro, R.~P. and Moniz, N. (2020).
\newblock Imbalanced regression and extreme value prediction.
\newblock {\em Mach. Learn.}, 109(9-10):1803--1835.

\bibitem[Rijsbergen, 1979]{Rijsbergen1979}
Rijsbergen, C. J.~V. (1979).
\newblock {\em Information Retrieval}.
\newblock Butterworth-Heinemann, Newton, MA, USA, 2nd edition.

\bibitem[Rousseeuw and Hubert, 2011]{Rousseeuw2011}
Rousseeuw, P.~J. and Hubert, M. (2011).
\newblock Robust statistics for outlier detection.
\newblock {\em WIREs Data Mining and Knowledge Discovery}, 1(1):73--79.

\bibitem[Torgo et~al., 2015]{torgo2015resampling}
Torgo, L., Branco, P., Ribeiro, R.~P., and Pfahringer, B. (2015).
\newblock Resampling strategies for regression.
\newblock {\em Expert Systems}, 32(3):465--476.

\bibitem[Torgo and Ribeiro, 2007]{Torgo2007}
Torgo, L. and Ribeiro, R. (2007).
\newblock Utility-based regression.
\newblock In {\em Proceedings of the 11th European Conference on Principles and
  Practice of Knowledge Discovery in Databases}, ECMLPKDD'07, page 597–604,
  Berlin, Heidelberg. Springer-Verlag.

\bibitem[Zheng et~al., 2013]{zheng2013u}
Zheng, Y., Liu, F., and Hsieh, H.-P. (2013).
\newblock U-air: When urban air quality inference meets big data.
\newblock In {\em Proceedings of the 19th ACM SIGKDD international conference
  on Knowledge discovery and data mining}, pages 1436--1444.

\end{thebibliography}

\end{document}